# Activity Recognition with Moving Cameras and Few Training Examples: Applications for Detection of Autism-Related Headbanging

**Peter Washington**, Bioengineering, Stanford University, Stanford, California, United States
**Aaron Kline**, Pediatrics, Stanford University, Stanford, California, United States
**Onur Cezmi Mutlu**, Stanford University, Stanford, California, United States
**Emilie Leblanc**, Pediatrics, Stanford University, Stanford, California, United States
**Cathy Hou**, Computer Science, Stanford University, Stanford, California, United States
**Nate Stockham**, Neuroscience, Stanford University, Stanford, California, United States
**Kelley Paskov**, Biomedical Data Science, Stanford University, Stanford, California, United States
**Brianna Chrisman**, Bioengineering, Stanford University, Stanford, California, United States
**Dennis P. Wall**, Pediatrics and Biomedical Data Science, Stanford University, Stanford, California, United States

Activity recognition computer vision algorithms can be used to detect the presence of autism-related behaviors, including what are termed "restricted and repetitive behaviors", or stimming, by diagnostic instruments. The limited data that exist in this domain are usually recorded with a handheld camera which can be shaky or even moving, posing a challenge for traditional feature representation approaches for activity detection which mistakenly capture the camera's motion as a feature. To address these issues, we first document the advantages and limitations of current feature representation techniques for activity recognition when applied to head banging detection. We then propose a feature representation consisting exclusively of head pose keypoints. We create a computer vision classifier for detecting head banging in home videos using a time-distributed convolutional neural network (CNN) in which a single CNN extracts features from each frame in the input sequence, and these extracted features are fed as input to a long short-term memory (LSTM) network. On the binary task of predicting head banging and no head banging within videos from the Self Stimulatory Behaviour Dataset (SSBD), we reach a mean F1-score of 90.77% using 3-fold cross validation (with individual fold F1-scores of 83.3%, 89.0%, and 100.0%) when ensuring that no child who appeared in the train set was in the test set for all folds. This work documents a successful technique for training a computer vision classifier which can detect human motion with few training examples and even when the camera recording the source clips is unstable. The general methods described here can be applied by designers and developers of interactive systems towards other human motion and pose classification problems used in mobile and ubiquitous interactive systems.

CCS CONCEPTS • Human-centered computing • Computing methodologies~Machine learning~Learning paradigms~Supervised learning • Computing methodologies~Artificial intelligence~Computer vision~Computer vision tasks~Activity recognition and understanding

## 1 INTRODUCTION

The increasing ubiquity of mobile devices in conjunction with advances in artificial intelligence (AI), and in particular machine learning (ML), are enabling accessibility of healthcare solutions for traditionally underserved populations. While the prevalence of autism exceeds 1 in 60 children [18, 29], current estimates are that over 80% of counties in the United States lack access to autism diagnostics [32]. As a result, families will often need to wait over a year to receive a diagnosis for autism, particularly in rural areas and in lower socioeconomic communities [19]. This prolonged and excessive wait time is particularly problematic for pediatric developmental delays like autism, where earlier diagnosis and resulting behavioral interventions result in more effective clinical outcomes [11].

Interactive devices such as smartphones are increasingly being evaluated for use in at-home digital diagnostics and therapies which adapt to the needs of the child [31]. For diagnostics, computer vision classifiers measure the behavior of the child in real-time, and statistical measurements are used to provide a diagnosis or screening [10, 43, 55]. For digital therapeutics, the classifiers measure the response of the child to an interactive game designed to provide behavioral therapy for the child [8-9, 24, 41, 42, 50].

Unfortunately, autism-related behaviors are currently particularly difficult to classify with computer vision. Existing activity recognition classifiers have been developed for common activities like running, throwing a basketball, and dancing [6]. Restricted and repetitive autism movements (i.e., stimming), such as hand flapping, head banging, and rocking back and forth, have not been the focus of computer vision efforts, largely because of a dearth of training data for such domain specific behaviors. An additional issue is that while many activity recognition algorithms assume a stable and still camera, many home videos of children exhibiting stimming behaviors are recorded from an unstable handheld camera source. There is a need for computer vision methods which can distinguish and detect human activity where massive training sets do not exist and only relatively small amounts of data from potentially unstable camera sources are available.

In this paper, we develop a feature representation method for head banging detection. We compare several traditional feature representation methods for activity recognition, visualizing each representation for a head banging clip recorded with a handheld camera. Through discussing the advantages and disadvantages of each technique, we arrive at a pose-based representation of only head keypoints. We train a time-distributed convolutional neural network (CNN) which extracts features that are fed into a long short-term memory (LSTM) network. We trained this model to distinguish video clips containing "head banging" from "not headbanging" using a balanced set of videos, resulting in a 90.77% F1-score during 3-fold cross-validation. This paper provides an example of training a human activity classifier with few data points and with training data from unstable camera streams, and we believe this work will be of utility to engineers of ML-powered interactive systems centered around human behavior.

## 2 RELATED WORK

We discuss examples of both advances in computer vision techniques for detecting stimming behaviors and corresponding interactive systems which target stimming behaviors of children with developmental delays. Because this research area is plentiful, we focus our literature review particularly on autism and pediatrics literature.

### 2.1 Computer vision detection of stimming

Computer vision approaches can enable scalable and affordable "behavioral phenotyping" of children with autism [35]. Here, we discuss prior computer vision efforts for automatically identifying motor movements and stereotyped behaviors.

Motor control and movements related to autism include head posture, head movement, and grasping patterns. Researchers have hypothesized that computer vision algorithms can pick up on the generally accepted motor variations between children with autism and neurotypical children. Some studies operate under controlled laboratory conditions. In one such study, Dawson et al. found that toddlers with autism exhibit increased head movement compared to neurotypical controls by tracking facial landmarks [10]. In another lab-controlled study, Zunino et al. used a CNN which extracts features which are then fed into an LSTM to detect grasping actions, reaching an accuracy of 72% for subjects with autism and 77% for neurotypical subjects [55]. We note that we employed a similar neural network architecture in the presents study. Some research efforts have worked towards classification of autism-related motor movements from unstructured video clips. Vyas et al. use a 2D mask R-CNN to distinguish autism from neurotypical behavior in unstructured video clips from a private data set, reporting precision of 72% and recall of 92% [43].

Advances in computer vision detection require shared and public datasets for training and testing the model. The Self-Stimulatory Behavior Dataset (SSBD) [34] is a dataset of unstructured home videos of children performing one of three self-stimulatory behaviors: headbanging, spinning, and hand flapping. The authors of the dataset built a 3-way classifier distinguishing these 3 behaviors using Space Time Interest Points (STIP) [26] and reach a 50.7% classification accuracy using 5-fold cross validation and 47.3% accuracy using 10-fold cross-validation. We use the SSBD dataset in the present study, as it is the only public dataset, we are aware of containing several labeled examples of self-stimulatory behaviors.

### 2.2 Machine learning powered interactive systems for behavioral therapy

Several interactive systems have been developed to aid in at-home behavioral therapy of children with autism using activity recognition models. Bartoli et al. evaluated the potential of games centered around "motion-based" touchless interaction, where sensing devices such as cameras powered by computer vision algorithms track and analyze body movements [3]. On a prototype of this paradigm with a Kinect application, they found that children with autism improved on measures of selective attention and sustained attention after playing the game. Robotic therapies powered by computer vision have also demonstrated great promise. Feil-Seifer et al. show that a classifier can determine whether children are attempting to interact with a socially assistive robot in a social manner [17]. Moghadas et al. used computer vision to distinguish children autism from neurotypical children from interaction with a parrot-like robot [30]. Continued improvements in the underlying computer vision algorithms powering these interactive systems will enable higher precision interactions in targeted therapeutic games for autism.

## 3  ACTIVITY-BASED FEATURE EXTRACTION WITH MOVING CAMERAS

In the absence of a massive and sufficiently heterogeneous dataset, activity recognition algorithms from video streams require feature extraction of relevant features to avoid overfitting to the training data. Before building our classifier, we explored three common activity recognition methods: Lucas-Kanade optical flow, dense optical flow, and pose estimation via keypoint detection. We use videos contained in the Self-Stimulatory Behavior Dataset (SSBD) [34], a dataset of unstructured home videos of children performing one of three self-stimulatory behaviors: headbanging, spinning, and hand flapping. The dataset was scraped from public domain videos posted on video sharing websites such as YouTube, Vimeo, and Dailymotion. Each of the three behaviors is marked with a timestamp. We manually visualized Lucas-Kanade optical flow, dense optical flow, and pose estimation via keypoint detection for several clips and determined that only a modified version of pose estimation would be feasible with moving cameras. Figure 1 displays the representation of 6 sequential frames using each technique for a video clip exhibiting head banging. We describe the advantages and limits of these feature representation methods below to motivate our eventual approach and use Figure 1 as an illustration of problems which occurred consistently across videos.

Dense optical flow computes flow for all points in a frame, resulting in "flow vectors" with a magnitude and direction [16]. Figure 1B displays an example of dense optical flow on a sequence of frames for a video clip of head banging in the SSBD dataset, where direction is encoded as hue and magnitude is encoded as intensity. Clear limitations of dense optical flow representations are immediately clear. For example, movement is detected outside of the region of the child, for example by the pink toy on the table. The child's chair is also sometimes detected as movement. These extra non-relevant movement patterns add much noise to the dense optical flow image, making it a non-ideal representation. In addition, the flow image appears similar to the outline of the child, making the method prone to overfitting to the child's body shape and camera angle.

Lucas-Kanade optical flow, in contrast to dense optical flow, computes flow for a sparse number of points pre-defined by the user [2], for example detected edges or corners. We create a uniform square grid of tracking points where each point is spaced out by 10 pixels in both the x and y direction. Figure 1C displays the resulting Lucas-Kanade optical flow motion tracking appended to the original image, while Figure 1D displays the tracking in isolation. The Lucas-Kanade method brings its own set of limitations by detecting movement outside of the body to an extent that is more dramatic than dense optical flow. This technique is particularly sensitive to background movement detection associated with slight shifts in the camera.

Pose estimation is a technique which is more robust to camera movement compared to optical flow. We use OpenPose realtime multi-person pose estimation [4-5, 36, 51] to track skeletal keypoints in each frame. OpenPose uses a CNN which is trained to predict part affinity fields, or flow fields representing relationships between body parts, and confidence maps which encode body part locations. Unlike optical flow, OpenPose predicts each frame independently of the surrounding frames. Figure 1E displays the estimated skeletal pose appended to the original image, while Figure 1D displays the pose skeleton in isolation. There are some clear limitations to using unmodified pose estimation. The noisy skeleton problem is a documented issue which arises when body parts are self-occluded [28, 54]. We observed that body part occlusion is a frequent occurrence in unstructured home videos, making unmodified pose estimation a non-ideal feature representation.

To account for the body part occlusion issue, which would inject unnecessary noise which would confuse the classifier, we modified the pose estimation output by only extracting key points in the head region. For a headbanging classifier that assumes the child subject is standing still, we hypothesized that only the head keypoints would be necessary for a headbanging classifier. The resulting feature representation is displayed in Figure 2, which contains 3 examples of headbanging sequences (A-C) and 3 examples of non-headbanging sequences (D-F). We qualitatively observed that the headbanging sequences tended to show differing keypoint positions for each frame, while non-head banging clips showed comparatively stable keypoints between frames (Figure 2).

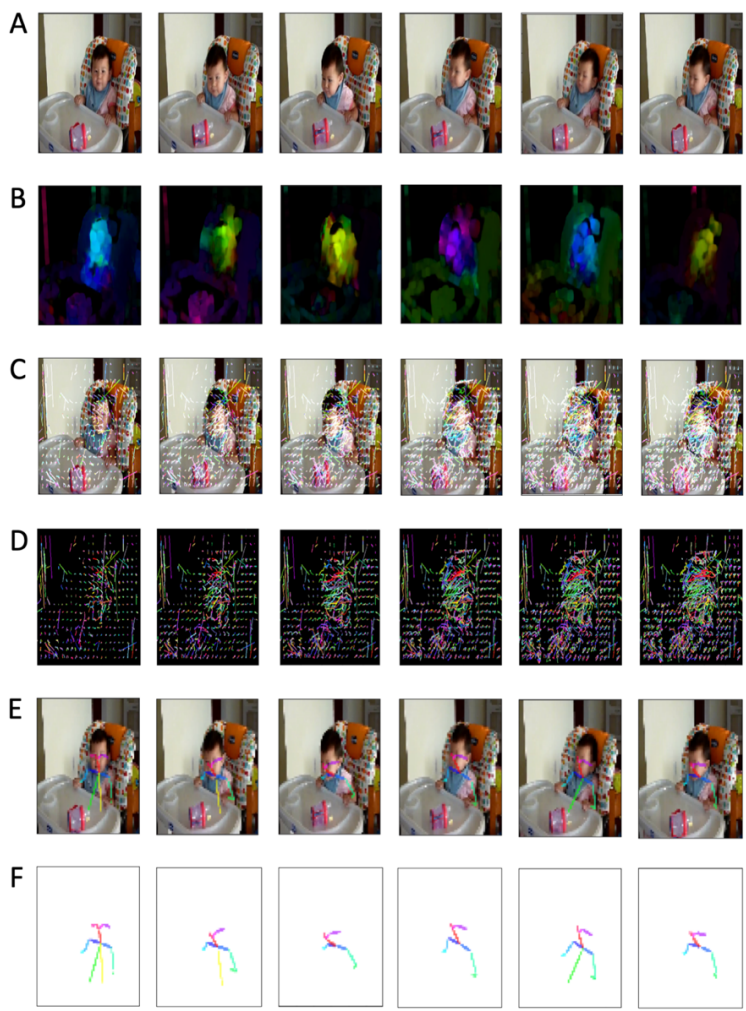

Figure 1: Representation of 6 sequential frames using popular feature representation techniques for activity recognition for a video clip exhibiting head banging. (A) The unmodified sequential frames. (B) Dense optical flow. (C) Lucas-Kanade optical flow displayed on the original image. (D) Lucas-Kanade optical flow in isolation. (E) Pose estimation using body keypoints displayed on the original image. (F) Pose estimation without the original image in the background. All methods displayed here contain noisy features unrelated to headbanging, many of which occur due to camera movement.

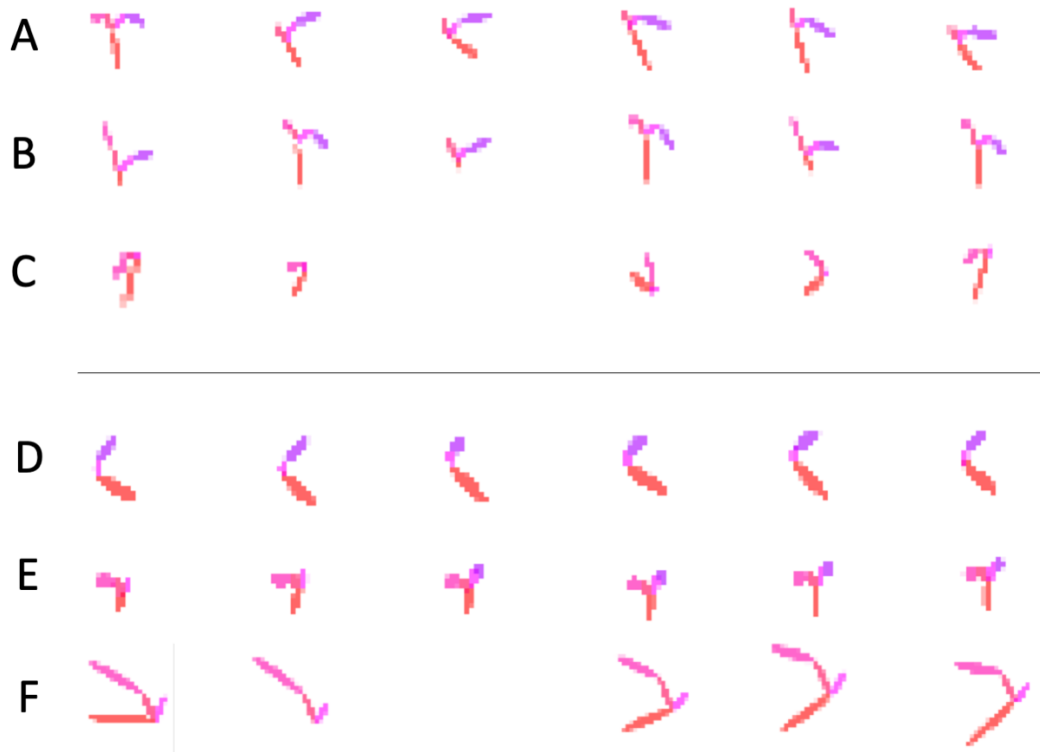

Figure 2: The feature representation used in this paper on headbanging video clips for 3 separate children (A) – (C) and for not head banging video clips for 3 other children (D) – (F). Only head keypoints and lines connecting them are retained in this representation. Every 5 frames in the original video are sampled to generate the input sequence to the neural network (7 frames covers one second of time at 30 frames per second). While head banging clips show differing keypoints for each frame, non-head banging clips show comparatively stable keypoints between frames.

## 4 TRAINING AND TESTING A HEAD BANGING DETECTOR

We implemented a time-distributed convolutional neural network (CNN) using the Keras [7] Python library with a Tensorflow [1] backend. The time-distributed CNN is a standard CNN architecture which learns to extract visual features from each frame in the image sequence. The activated feature maps from each image in the video clip are fed into the corresponding sequence input position in a long short-term memory (LSTM) [22] neural network. We note that the same CNN architecture and weights are used at each time step in the LSTM rather than a separate CNN trained for each position in the sequence. Figure 3 provides a visualization of the time-distributed CNN architecture. We trained using Adam optimization [23] with an initial learning rate of 0.0001.

We performed 3-fold cross validation, ensuring that no child who appeared in the train set would appear in the test set for all folds. Three folds was the right amount to ensure that a variety of children would appear in both the training and testing sets while still maintaining the ability to evaluate the classifier under multiple partitions of the dataset. In total, we used 27 video clips containing headbanging video clips containing "normal" head motions. To minimize overfitting and increase generalization, we applied the following data augmentations to each frame: rotation at a random interval between -45 and 45 degrees and zooming in with a random zoom factor between 1.0 and 2.0. Although the dataset is small, we believe that the combination of the reduced feature space, data augmentation, and complete separation of children between training and testing sets provides strong validation of the presented technique for detecting head banging.

The mean F1-score, a popular machine learning metric which is the harmonic mean of precision and recall, was 90.77% across the 3 cross-validation folds. The individual F1-scores per fold were 83.3%, 89.0%, and 100.0%.

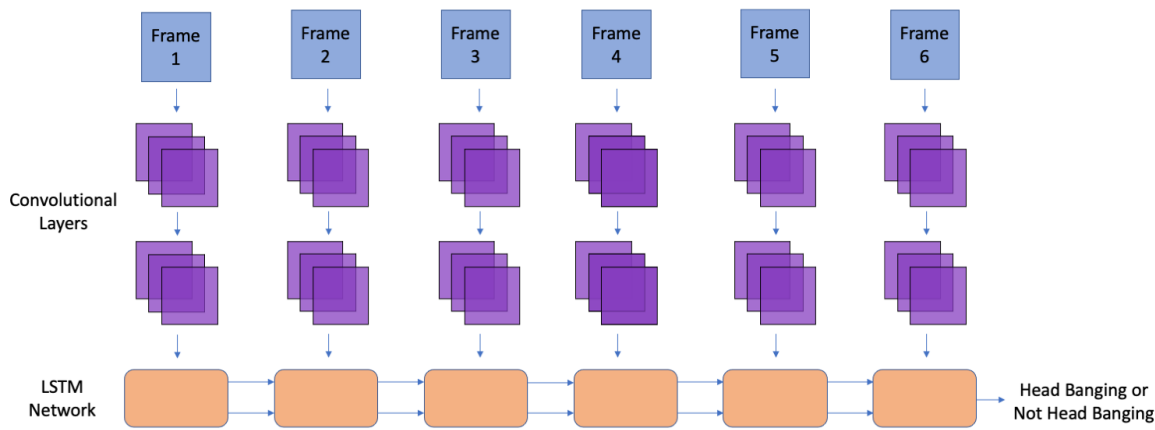

Figure 3: Model architecture for the time-distributed convolutional neural network (CNN) trained to detect hand flapping from short video sequences. Each input frame is fed into a CNN which extracts visual features from the frame. These features are used as input to an LSTM network which makes the final binary prediction of "head banging" vs. "not head banging".

## 5   DISCUSSION AND FUTURE WORK

We presented a computer vision classifier which can detect head banging in videos recorded with a handheld device. We achieve a 90.77% F1-score on video clips from the SSBD dataset. We separated videos by children into the train and test sets for each fold of 3-fold cross validation, ensuring that no child in a video clip which appeared in the train set also appeared in the test set. However, we must admit that head banging classification is an incredibly "easy" computer vision task after extracting the correct feature representation of exclusively head poses, as visualized in Figure 3. This highlights the importance of selecting a robust and representative feature representation when conducting activity-based activity recognition. Other autism-related stimming behaviors, such as hand flapping and body spinning, will likely require further engineering of the feature representation.

A major direction for future work will be to evaluate the utility of the head banging model in clinical settings. Remote AI-powered video-based diagnostics and screeners [13-15, 21, 25, 27, 39-40, 44-48] as well as telemedical digital therapies [8-9, 12, 20, 33, 37-38, 41-42, 49-50, 52-53] are increasingly being developed for autism in children. The efficacy of remote and automated telemedicine is limited to the performance of the underlying AI. Due to the complexity of computer vision for human behavior in general, further innovations and creativity in feature extraction and representation methods are vital to account for noisy datasets and to enable distinguishing of complex human behaviors. Crucially, at-home therapy requires more robust vision models which can account for moving or unsteady cameras, which are conditions that are not usually accounted for by optical flow and other feature representations for activity recognition. Extracting a subset of poses is a good start, but there are potentially more effective and robust representations waiting to be discovered.

There are several HCI considerations which warrant future study. It would be useful to know whether crowd workers can provide reliable activity annotations with transformed images to enables privacy-aware crowdsourcing. For example, while optical flow may not be an effective representation for training ML models, it may preserve the privacy of the clips while still revealing enough information to allow crowd workers to provide correct labels. It will also be important to evaluate how new behavioral classifiers can be integrated into interactive systems, enabling increasingly adaptive user interfaces.

## 6 CONCLUSION

We presented a technique of utility to interaction designers and engineers of interactive systems that measure human motion patterns. We consider the reality that human activity is often tracked with moving cameras, posing an issue for several traditional feature representations such as optical flow or using full body keypoints. While we validated our method for head banging detection, a problem that is especially crucial for digital autism diagnostics and adaptive therapeutics, the presented technique can be applied by practitioners of interactive and AI-powered systems which require measurement of human motion under unstable camera conditions.

**ACKNOWLEDGMENTS**

This work was supported in part by funds to DPW from the National Institutes of Health (1R01EB025025-01, 1R21HD091500-01, 1R01LM013083), the National Science Foundation (Award 2014232), The Hartwell Foundation, Bill and Melinda Gates Foundation, Coulter Foundation, Lucile Packard Foundation, the Weston Havens Foundation, and program grants from Stanford's Human Centered Artificial Intelligence Program, Stanford's Precision Health and Integrated Diagnostics Center (PHIND), Stanford's Beckman Center, Stanford's Bio-X Center, Predictives and Diagnostics Accelerator (SPADA) Spectrum, Stanford's Spark Program in Translational Research, and from Stanford's Wu Tsai Neurosciences Institute's Neuroscience: Translate Program. We also acknowledge generous support from David Orr, Imma Calvo, Bobby Dekesyer and Peter Sullivan. P.W. would like to acknowledge support from Mr. Schroeder and the Stanford Interdisciplinary Graduate Fellowship (SIGF) as the Schroeder Family Goldman Sachs Graduate Fellow.

**REFERENCES**

[1] Abadi, Martín, Paul Barham, Jianmin Chen, Zhifeng Chen, Andy Davis, Jeffrey Dean, Matthieu Devin et al. "Tensorflow: A system for large-scale machine learning." In *12th {USENIX} symposium on operating systems design and implementation ({OSDI} 16)*, pp. 265-283. 2016.

[2] Baker, Simon, and Iain Matthews. "Lucas-kanade 20 years on: A unifying framework." *International journal of computer vision* 56, no. 3 (2004): 221-255.

[3] Bartoli, Laura, Clara Corradi, Franca Garzotto, and Matteo Valoriani. "Exploring motion-based touchless games for autistic children's learning." In *Proceedings of the 12th international conference on interaction design and children*, pp. 102-111. 2013.

[4] Cao, Zhe, Gines Hidalgo, Tomas Simon, Shih-En Wei, and Yaser Sheikh. "OpenPose: realtime multi-person 2D pose estimation using Part Affinity Fields." *IEEE transactions on pattern analysis and machine intelligence* 43, no. 1 (2019): 172-186.

[5] Cao, Zhe, Tomas Simon, Shih-En Wei, and Yaser Sheikh. "Realtime multi-person 2d pose estimation using part affinity fields."

In *Proceedings of the IEEE conference on computer vision and pattern recognition*, pp. 7291-7299. 2017.

[6] Caba Heilbron, Fabian, Victor Escorcia, Bernard Ghanem, and Juan Carlos Niebles. "Activitynet: A large-scale video benchmark for human activity understanding." In *Proceedings of the ieee conference on computer vision and pattern recognition*, pp. 961-970. 2015.

[7] Chollet, François. "Keras: The python deep learning library." *ascl* (2018): ascl-1806.

[8] Daniels, Jena, Jessey N. Schwartz, Catalin Voss, Nick Haber, Azar Fazel, Aaron Kline, Peter Washington, Carl Feinstein, Terry Winograd, and Dennis P. Wall. "Exploratory study examining the at-home feasibility of a wearable tool for social-affective learning in children with autism." *NPJ digital medicine* 1, no. 1 (2018): 1-10.

[9] Daniels, Jena, Nick Haber, Catalin Voss, Jessey Schwartz, Serena Tamura, Azar Fazel, Aaron Kline et al. "Feasibility testing of a wearable behavioral aid for social learning in children with autism." *Applied clinical informatics* 9, no. 1 (2018): 129.

[10] Dawson, Geraldine, Kathleen Campbell, Jordan Hashemi, Steven J. Lippmann, Valerie Smith, Kimberly Carpenter, Helen Egger et al. "Atypical postural control can be detected via computer vision analysis in toddlers with autism spectrum disorder." *Scientific reports* 8, no. 1 (2018): 1-7.

[11] Dawson, Geraldine, and Guillermo Sapiro. "Potential for digital behavioral measurement tools to transform the detection and diagnosis of autism spectrum disorder." *JAMA pediatrics* 173, no. 4 (2019): 305-306.

[12] Deriso, David, Joshua Susskind, Lauren Krieger, and Marian Bartlett. "Emotion mirror: a novel intervention for autism based on real-time expression recognition." In *European Conference on Computer Vision*, pp. 671-674. Springer, Berlin, Heidelberg, 2012.

[13] Duda, Marlena, Jena Daniels, and Dennis P. Wall. "Clinical evaluation of a novel and mobile autism risk assessment." *Journal of autism and developmental disorders* 46, no. 6 (2016): 1953-1961.

[14] Duda, M., N. Haber, J. Daniels, and D. P. Wall. "Crowdsourced validation of a machine-learning classification system for autism and ADHD." *Translational psychiatry* 7, no. 5 (2017): e1133-e1133.

[15] Duda, M., R. Ma, N. Haber, and D. P. Wall. "Use of machine learning for behavioral distinction of autism and ADHD." *Translational psychiatry* 6, no. 2 (2016): e732-e732.

[16] Farnebäck, Gunnar. "Two-frame motion estimation based on polynomial expansion." In *Scandinavian conference on Image analysis*, pp. 363-370. Springer, Berlin, Heidelberg, 2003.

[17] Feil-Seifer, David, and Maja Matarić. "Using proxemics to evaluate human-robot interaction." In *2010 5th ACM/IEEE International Conference on Human-Robot Interaction (HRI)*, pp. 143-144. IEEE, 2010.

[18] Fombonne, Eric. "The rising prevalence of autism." *Journal of Child Psychology and Psychiatry* 59, no. 7 (2018): 717-720.

[19] Gordon-Lipkin, Eliza, Jessica Foster, and Georgina Peacock. "Whittling down the wait time: exploring models to minimize the delay from initial concern to diagnosis and treatment of autism spectrum disorder." *Pediatric Clinics* 63, no. 5 (2016): 851-859.

[20] Haber, Nick, Catalin Voss, and Dennis Wall. "Making emotions transparent: Google Glass helps autistic kids understand facial expressions through augmented-reaiity therapy." *IEEE Spectrum* 57, no. 4 (2020): 46-52.

[21] Halim, Abbas, Garberson Ford, Stuart Liu-Mayo, Eric Glover, and Dennis P. Wall. "Multi-modular AI Approach to Streamline Autism Diagnosis in Young Children." *Scientific Reports (Nature Publisher Group)* 10, no. 1 (2020).

[22] Hochreiter, Sepp, and Jürgen Schmidhuber. "Long short-term memory. " Neural computation 9, no. 8 (1997): 1735-1780.

[23] Kingma, Diederik P., and Jimmy Ba. "Adam: A method for stochastic optimization." *arXiv preprint arXiv:1412.6980* (2014).

[24] Kline, Aaron, Catalin Voss, Peter Washington, Nick Haber, Hessey Schwartz, Qandeel Tariq, Terry Winograd, Carl Feinstein, and Dennis P. Wall. "Superpower glass." *GetMobile: Mobile Computing and Communications* 23, no. 2 (2019): 35-38.

[25] Kosmicki, J. A., V. Sochat, M. Duda, and D. P. Wall. "Searching for a minimal set of behaviors for autism detection through feature selection-based machine learning." *Translational psychiatry* 5, no. 2 (2015): e514-e514.

[26] Laptev, Ivan. "On space-time interest points." *International journal of computer vision* 64, no. 2-3 (2005): 107-123.

[27] Leblanc, Emilie, Peter Washington, Maya Varma, Kaitlyn Dunlap, Yordan Penev, Aaron Kline, and Dennis P. Wall. "Feature replacement methods enable reliable home video analysis for machine learning detection of autism." *Scientific reports* 10, no. 1 (2020): 1-11.

[28] Li, Wanqing, Zhengyou Zhang, and Zicheng Liu. "Action recognition based on a bag of 3d points." In *2010 IEEE Computer Society Conference on Computer Vision and Pattern Recognition-Workshops*, pp. 9-14. IEEE, 2010.

[29] Matson, Johnny L., and Alison M. Kozlowski. "The increasing prevalence of autism spectrum disorders." *Research in Autism Spectrum Disorders* 5, no. 1 (2011): 418-425.

[30] Moghadas, M., and H. Moradi. "Analyzing Human-Robot Interaction Using Machine Vision for Autism screening." In *2018 6th RSI International Conference on Robotics and Mechatronics (IcRoM)*, pp. 572-576. IEEE, 2018.

[31] Moon, Sun Jae, Jinseub Hwang, Harrison Scott Hill, Ryan Kervin, Kirstin Brown Birtwell, John Torous, Christopher J. McDougle, and Jung Won Kim. "Mobile device applications and treatment of autism spectrum disorder: a systematic review and meta-analysis of effectiveness." *Archives of Disease in Childhood* 105, no. 5 (2020): 458-462.

[32] Ning, Michael, Jena Daniels, Jessey Schwartz, Kaitlyn Dunlap, Peter Washington, Haik Kalantarian, Michael Du, and Dennis P. Wall. "Identification and quantification of gaps in access to autism resources in the United States: an infodemiological study." *Journal of Medical Internet Research* 21, no. 7 (2019): e13094.

[33] Pioggia, Giovanni, Roberta Igliozzi, Marcello Ferro, Arti Ahluwalia, Filippo Muratori, and Danilo De Rossi. "An android for enhancing social skills and emotion recognition in people with autism." *IEEE Transactions on Neural Systems and Rehabilitation Engineering* 13, no. 4

(2005): 507-515.

[34] Rajagopalan, Shyam, Abhinav Dhall, and Roland Goecke. "Self-stimulatory behaviours in the wild for autism diagnosis." In *Proceedings of the IEEE International Conference on Computer Vision Workshops*, pp. 755-761. 2013.

[35] Sapiro, Guillermo, Jordan Hashemi, and Geraldine Dawson. "Computer vision and behavioral phenotyping: an autism case study." *Current Opinion in Biomedical Engineering* 9 (2019): 14-20.

[36] Simon, Tomas, Hanbyul Joo, Iain Matthews, and Yaser Sheikh. "Hand keypoint detection in single images using multiview bootstrapping." In *Proceedings of the IEEE conference on Computer Vision and Pattern Recognition*, pp. 1145-1153. 2017.

[37] Slovák, Petr, Ran Gilad-Bachrach, and Geraldine Fitzpatrick. "Designing social and emotional skills training: The challenges and opportunities for technology support." In *Proceedings of the 33rd Annual ACM Conference on Human Factors in Computing Systems*, pp. 2797-2800. 2015.

[38] Smitha, Kavallur Gopi, and A. Prasad Vinod. "Facial emotion recognition system for autistic children: a feasible study based on FPGA implementation." *Medical & biological engineering & computing* 53, no. 11 (2015): 1221-1229.

[39] Tariq, Qandeel, Scott Lanyon Fleming, Jessey Nicole Schwartz, Kaitlyn Dunlap, Conor Corbin, Peter Washington, Haik Kalantarian, Naila Z. Khan, Gary L. Darmstadt, and Dennis Paul Wall. "Detecting developmental delay and autism through machine learning models using home videos of Bangladeshi children: Development and validation study." *Journal of medical Internet research* 21, no. 4 (2019): e13822.

[40] Tariq, Qandeel, Jena Daniels, Jessey Nicole Schwartz, Peter Washington, Haik Kalantarian, and Dennis Paul Wall. "Mobile detection of autism through machine learning on home video: A development and prospective validation study." *PLoS medicine* 15, no. 11 (2018): e1002705.

[41] Voss, Catalin, Jessey Schwartz, Jena Daniels, Aaron Kline, Nick Haber, Peter Washington, Qandeel Tariq et al. "Effect of wearable digital intervention for improving socialization in children with autism spectrum disorder: a randomized clinical trial." *JAMA pediatrics* 173, no. 5 (2019): 446-454.

[42] Voss, Catalin, Peter Washington, Nick Haber, Aaron Kline, Jena Daniels, Azar Fazel, Titas De et al. "Superpower glass: delivering unobtrusive real-time social cues in wearable systems." In *Proceedings of the 2016 ACM International Joint Conference on Pervasive and Ubiquitous Computing: Adjunct*, pp. 1218-1226. 2016.

[43] Vyas, Kathan, Rui Ma, Behnaz Rezaei, Shuangjun Liu, Michael Neubauer, Thomas Ploetz, Ronald Oberleitner, and Sarah Ostadabbas. "Recognition Of Atypical Behavior In Autism Diagnosis From Video Using Pose Estimation Over Time." In *2019 IEEE 29th International Workshop on Machine Learning for Signal Processing (MLSP)*, pp. 1-6. IEEE, 2019.

[44] Wall, Dennis Paul, J. Kosmicki, T. F. Deluca, E. Harstad, and Vincent Alfred Fusaro. "Use of machine learning to shorten observation-based screening and diagnosis of autism." *Translational psychiatry* 2, no. 4 (2012): e100-e100.

[45] Washington, Peter, Emilie Leblanc, Kaitlyn Dunlap, Yordan Penev, Aaron Kline, Kelley Paskov, Min Woo Sun et al. "Precision Telemedicine through Crowdsourced Machine Learning: Testing Variability of Crowd Workers for Video-Based Autism Feature Recognition." *Journal of personalized medicine* 10, no. 3 (2020): 86.

[46] Washington, Peter, Natalie Park, Parishkrita Srivastava, Catalin Voss, Aaron Kline, Maya Varma, Qandeel Tariq et al. "Data-driven diagnostics and the potential of mobile artificial intelligence for digital therapeutic phenotyping in computational psychiatry." *Biological Psychiatry: Cognitive Neuroscience and Neuroimaging* (2019).

[47] Washington, Peter, Haik Kalantarian, Qandeel Tariq, Jessey Schwartz, Kaitlyn Dunlap, Brianna Chrisman, Maya Varma et al. "Validity of online screening for autism: crowdsourcing study comparing paid and unpaid diagnostic tasks." *Journal of medical Internet research* 21, no. 5 (2019): e13668.

[48] Washington, Peter, Kelley Marie Paskov, Haik Kalantarian, Nathaniel Stockham, Catalin Voss, Aaron Kline, Ritik Patnaik et al. "Feature selection and dimension reduction of social autism data." In *Pac Symp Biocomput*, vol. 25, pp. 707-718. 2020.

[49] Washington, Peter, Catalin Voss, Nick Haber, Serena Tanaka, Jena Daniels, Carl Feinstein, Terry Winograd, and Dennis Wall. "A wearable social interaction aid for children with autism." In *Proceedings of the 2016 CHI Conference Extended Abstracts on Human Factors in Computing Systems*, pp. 2348-2354. 2016.

[50] Washington, Peter, Catalin Voss, Aaron Kline, Nick Haber, Jena Daniels, Azar Fazel, Titas De, Carl Feinstein, Terry Winograd, and Dennis Wall. "SuperpowerGlass: a wearable aid for the at-home therapy of children with autism." *Proceedings of the ACM on interactive, mobile, wearable and ubiquitous technologies* 1, no. 3 (2017): 1-22.

[51] Wei, Shih-En, Varun Ramakrishna, Takeo Kanade, and Yaser Sheikh. "Convolutional pose machines." In *Proceedings of the IEEE conference on Computer Vision and Pattern Recognition*, pp. 4724-4732. 2016.

[52] White, Susan W., Lynn Abbott, Andrea Trubanova Wieckowski, Nicole N. Capriola-Hall, Sherin Aly, and Amira Youssef. "Feasibility of automated training for facial emotion expression and recognition in autism." *Behavior therapy* 49, no. 6 (2018): 881-888.

[53] Wright, Peter, and John McCarthy. "Empathy and experience in HCI." In *Proceedings of the SIGCHI conference on human factors in computing systems*, pp. 637-646. 2008.

[54] Zhang, Shugang, Zhiqiang Wei, Jie Nie, Lei Huang, Shuang Wang, and Zhen Li. "A review on human activity recognition using vision-based method." *Journal of healthcare engineering* 2017 (2017).

[55] Zunino, Andrea, Pietro Morerio, Andrea Cavallo, Caterina Ansuini, Jessica Podda, Francesca Battaglia, Edvige Veneselli, Cristina Becchio, and Vittorio Murino. "Video gesture analysis for autism spectrum disorder detection." In *2018 24th International Conference on Pattern Recognition (ICPR)*, pp. 3421-3426. IEEE, 2018.